\documentclass[sigconf]{acmart}

\AtBeginDocument{%
  }

\setcopyright{acmlicensed}
\copyrightyear{2018}
\acmYear{2018}
\acmDOI{XXXXXXX.XXXXXXX}

\acmConference[Conference acronym 'XX]{Make sure to enter the correct
  conference title from your rights confirmation emai}{June 03--05,
  2018}{Woodstock, NY}
\acmISBN{978-1-4503-XXXX-X/18/06}

\begin{document}

\title{An Effective Approach for Node Classification in Textual Graphs}

\author{Rituparna Datta$^*$}
\email{hht9zt@virginia.edu}
\affiliation{%
  \institution{University of Virginia}
  \state{Virginia}
  \country{USA}
}
\author{Nibir Chandra Mandal$^*$}
\email{wyr6fx@virginia.edu}
\affiliation{%
  \institution{University of Virginia}
  \state{Virginia}
  \country{USA}
}








\renewcommand{\shortauthors}{Trovato et al.}

\maketitle
\def\thefootnote{*}\footnotetext{These authors contributed equally to this work}\def\thefootnote{\arabic{footnote}}
\section{Introduction}
In modern knowledge systems, Textual Attribute Graphs (TAGs) integrate textual metadata with graph connectivity to model complex networks such as citation networks, recommendation systems, and knowledge graphs. These TAGs hold immense potential for tasks like node classification, where the goal is to predict labels (e.g., topics or categories) for graph nodes based on structural and textual data. Despite their promise, existing methods often fail to fully utilize the intricate relationships within TAGs due to limitations in capturing semantic richness from text or modeling long-range dependencies in graph structures. Recent advances in large language models (LLMs) and graph neural networks (GNNs) have inspired hybrid solutions, but challenges persist in scalability, integration, and dynamic adaptability.

Citation networks represent a prime example and critical application of TAGs, forming the backbone of scientific knowledge organization. In these networks, academic papers serve as nodes connected through citation relationships, with each paper containing rich textual content in the form of titles, abstracts, and full text. The accurate classification of scientific papers is crucial for several reasons: ($i$) scientific repositories containing millions of papers require automated methods for organizing and categorizing research; ($ii$) effective classification enables researchers to discover relevant work across disciplinary boundaries; ($iii$) understanding paper categories helps track the evolution of scientific fields and emerging research areas; ($iv$) funding agencies and institutions need accurate classification for research impact assessment and resource distribution.

In the context of citation networks and broader TAG applications, several interconnected challenges hinder effective node classification as follows:
\begin{itemize}
    \item \textbf{Semantic-Structural Integration}: Scientific papers contain complex domain-specific terminology and concepts that require sophisticated semantic understanding. While traditional text embeddings can capture basic meaning, they often fail to represent the nuanced technical content. Additionally, citation patterns form intricate hierarchical relationships that carry important structural information. Current approaches either process these aspects separately or combine them through simple concatenation, failing to capture the rich interplay between semantic content and citation structure.
    \item \textbf{Long-range Dependencies}: Scientific influence often propagates through multiple layers of citations, creating important relationships between papers that are several citation hops apart. Traditional GNN architectures, limited by their local aggregation mechanisms, struggle to capture these distant relationships effectively. This limitation is particularly problematic in citation networks where papers can significantly influence a field through chains of intermediate citations, forming patterns that local graph structures cannot adequately represent.

    \item \textbf{Temporal Evolution}: Citation networks are inherently dynamic, with both research topics and citation patterns evolving over time. New research areas emerge, terminology shifts, and citation behaviors change, making it challenging for models trained on historical data to maintain performance on newer papers. This temporal aspect is explicitly demonstrated in the ogbn-arxiv dataset, where the training set consists of older papers while the test set contains more recent ones, directly challenging models' ability to generalize across time periods.

    \item \textbf{Scalability}: Contemporary citation networks encompass millions of papers with complex interconnections, presenting significant computational challenges. Processing both detailed textual content and extensive graph structures requires substantial computational resources. The challenge is further amplified by the need for efficient real-time inference when classifying new papers, making it crucial to balance model sophistication with computational efficiency.
\end{itemize}

To address these interconnected challenges, we propose a framework that integrates TAPE (Text-Attributed Graph Representation Enhancement) with Graphormer, creating a comprehensive solution for node classification in citation networks. At the semantic-structural level, TAPE employs ChatGPT to generate rich semantic explanations from paper content, which are then processed by a lightweight transformer into dense embeddings and combined with structural information through attention mechanisms. The framework handles long-range dependencies through Graphormer's attention mechanism, which directly models relationships between distant nodes using path-aware position encoding and multi-head attention, enabling the model to capture various types of long-range relationships. For temporal evolution, TAPE generates semantic representations focused on core scientific concepts while abstracting away temporal-specific language, with a training procedure optimized for generalization to newer papers. The framework achieves scalability through an efficient two-stage architecture that separates semantic processing from structural learning, utilizing parallelizable attention mechanisms in Graphormer and implementing strategic caching of TAPE embeddings for frequently accessed papers.

Our work makes several significant technical contributions to the field of node classification in citation networks and TAGs more broadly. First, we introduce a novel architectural framework that uniquely integrates LLM-based semantic processing through TAPE with Graphormer's structural learning capabilities, featuring a new attention mechanism specifically designed for scientific paper classification and an efficient two-stage processing pipeline. Second, we enhance node representations through the TAPE framework, which generates four distinct types of embeddings: explanatory, predictive, textual, and graph-based features, combined through a novel integration layer with learned attention weights, resulting in empirically validated improvements in node representation quality. Third, our approach achieves state-of-the-art performance on the ogbn-arxiv dataset with 0.772 classification accuracy, significantly outperforming the previous best GCN baseline (0.713), while maintaining consistent improvements across precision (0.671), recall (0.577), and F1-score (0.610). Finally, we provide comprehensive empirical validation through detailed ablation studies quantifying individual component contributions.




Specifically we have the following contribution in this work:
\begin{itemize}
\item We propose a novel LLM-driven approach for node classification in text-attributed graphs, where ChatGPT generates semantic-rich explanations that are transformed into enhanced node representations. This approach specifically addresses the challenge of capturing complex domain-specific terminology in scientific papers by leveraging the advanced language understanding capabilities of LLMs.
\item We develop TAPE-Graphormer, an integrated framework that combines TAPE's semantic processing with Graphormer's structural learning. TAPE generates four complementary embeddings that are fused through learned attention weights, while Graphormer's path-aware position encoding and multi-head attention mechanisms capture complex citation relationships across multiple hops.

\item We conduct empirical evaluation on the ogbn-arxiv dataset, demonstrating improvements over existing methods: 0.772 classification accuracy (vs. 0.713 GCN baseline), 0.671 precision, 0.577 recall, and 0.610 F1-score.

\item We provide comprehensive analysis through ablation studies quantifying the impact of each component (TAPE embeddings, attention mechanisms, integration layer) on classification performance. Our results demonstrate that combining semantic and structural information through our approach yields consistent improvements across different research domains and citation patterns in large-scale networks containing millions of papers.
\end{itemize}

\section{Existing Work}

In the context of Textual Attribute Graphs (TAGs), traditional methods often involve the transformation of text attributes into shallow or hand-crafted features such as skip-gram \cite{mikolov2013efficient} or Bag-of-Words (BoW) \cite{harris1954distributional} embeddings. These shallow features are widely used in graph learning due to their simplicity and efficiency, particularly when fed into graph-based learning algorithms like Graph Convolutional Networks (GCNs) \cite{kipf2016semi}. Several works have explored shallow embeddings in the design of GNN architectures \cite{velickovic2017graph, chiang2019cluster, velickovic2019deep} and for benchmarking purposes in graph representation learning \cite{yang2016revisiting, hu2020open}. However, these methods may be limited in capturing complex semantic relationships and contextual information, making it difficult to fully leverage the richness of text attributes in intricate graph learning tasks.

To address the limitations of shallow embedding approaches, researchers have turned to deep embeddings using pre-trained Language Models (LMs) such as BERT \cite{devlin2018bert}. These methods integrate LMs and GNNs in various architectures, with some approaches cascading LM-based text encodings into GNNs, as seen in models like TextGNN \cite{zhu2021textgnn}, GIANT \cite{chien2021giant}, and GPT-GNN \cite{hu2020gpt}. Other works employ iterative workflows that simultaneously refine both text representations and node embeddings, exemplified by Graphormer \cite{ying2021transformers}, DRAGON \cite{yasunaga2022dragon}, and GLEM \cite{zhao2022gtn}. Recent explorations into incorporating large language models (LLMs) such as ChatGPT \cite{brown2020language}, PaLM \cite{chowdhery2022palm}, and LLaMA \cite{touvron2023llama} into TAG tasks further demonstrate their potential in advancing graph-based natural language processing.

\section{Problem Statement}

Given a directed citation network represented as a graph $G = (V, E)$, where $V = \{v_1, v_2, ..., v_n\}$ denotes the set of papers and $E \subseteq V \times V$ represents the citation relationships between papers. Each paper $v_i \in V$ is associated with textual content $T_i$, comprising its title and abstract, and belongs to a research category $y_i \in Y$, where $Y$ is the set of possible research domains. The node classification task aims to learn a function $f: (G, T) \rightarrow Y$ that maps the graph structure and textual content to research categories. This function must satisfy several key requirements: (1) effectively encode and utilize the semantic information present in the text content $T_i$ of each paper, (2) meaningfully integrate these textual features with the structural properties derived from the citation graph $G$, (3) generalize well to predict accurate research categories for previously unseen papers, particularly those published after the training set papers, and (4) maintain computational efficiency when scaling to large-scale citation networks containing millions of papers and citations. The temporal aspect is particularly important as the training set consists of papers from earlier time periods, while the model must make predictions on more recent papers in the test set.

\begin{figure*}[h]
    \centering
    \includegraphics[width=0.9\linewidth]{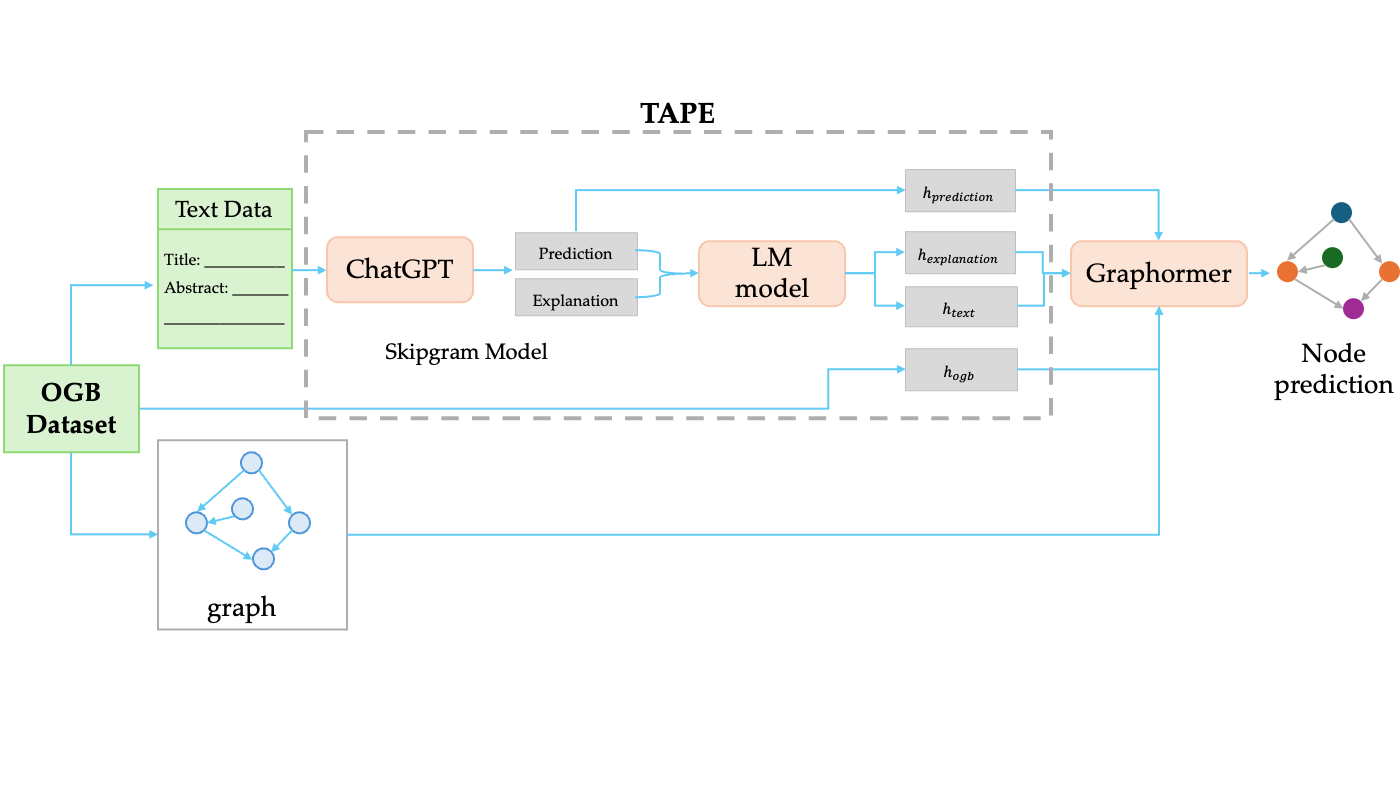}
    \vspace{-17mm}
    \caption{Proposed Model for Node Classification. This framework harnesses large language models (LLMs) to improve representation learning on textual attribute graphs (TAGs). At first, the textual attributes of each node, such as the \textit{title and abstract}, are formatted into a custom prompt (green box) to query the LLM, for our case ChatGPT, which produces a ranked \textit{prediction} list along with an \textit{explanation}. Subsequently, the original text, predictions, and explanations are employed to fine-tune a language model (LM), which is then transformed into vectorized node features. Ultimately, these enhanced node features—namely, \(h_{\text{expl}}\), \(h_{\text{pred}}\), \(h_{\text{text}}\) and  \(h_{\text{ogb}}\) are integrated into Graphormer for the prediction of unknown node classes.
    }
    
    \label{fig:enter-label}
\end{figure*}
\section{Proposed Methodology}
Our proposed framework combines TAPE-enhanced semantic representations with Graphormer's structural learning capabilities to effectively address the challenges in node classification for citation networks. Figure 1 illustrates the overall architecture of our approach, which we detail below.

\subsection{Input Processing}
\paragraph{Text Input} We leverage both titles and abstracts from papers as they provide complementary information: titles typically contain concise topic indicators, while abstracts offer detailed technical content and methodology descriptions. This dual-source approach ensures robust semantic understanding compared to title-only or abstract-only approaches.

\paragraph{Graph Input} We utilize the citation network structure represented as a directed graph $G = (V, E)$, where $V$ represents papers and $E$ represents citations. The graph structure is crucial for capturing scholarly influence patterns and research lineage that pure text-based approaches would miss.

\subsection{TAPE Framework}
Our TAPE framework processes textual content through three stages:

\paragraph{LLM Processing} We employ ChatGPT to generate semantic-rich explanations, leveraging its pre-trained knowledge of scientific concepts. This choice is motivated by LLMs' superior ability to understand technical terminology and context compared to traditional embedding methods. The LLM processes input text $T_i$ for each node $i$ to produce:
\begin{equation}
    h_{expl}, h_{pred} = \text{LLM}(T_i)
\end{equation}
where $h_{expl}$ represents explanation embeddings and $h_{pred}$ represents prediction embeddings.

\paragraph{Lightweight Transformation} We employ a transformer-based interpreter to convert LLM outputs into compact, fixed-dimensional representations. This step is critical for:
\begin{itemize}
    \item Dimensionality reduction while preserving semantic information
    \item Ensuring computational efficiency in downstream processing
    \item Standardizing representations across different text lengths
\end{itemize}

\paragraph{Feature Integration} We combine multiple embedding types:
\begin{equation}
    h_{final} = \text{Attention}([h_{expl}; h_{pred}; h_{text}; h_{ogb}])
\end{equation}
where $[;]$ denotes concatenation and $\text{Attention}(\cdot)$ is a learned attention mechanism.

\subsection{Graph Structure Encoding}
We encode the citation graph structure using two complementary components:

\paragraph{Positional Encoding} We implement path-aware positional encoding $PE(i,j)$ for nodes $i$ and $j$:
\begin{equation}
    PE(i,j) = f(d_{ij}, p_{ij})
\end{equation}
where $d_{ij}$ is the shortest path distance and $p_{ij}$ represents the path features. This encoding captures both local and global structural information.

\paragraph{Structural Features} We compute structural features including:
\begin{itemize}
    \item Node degree statistics (in/out degree)
    \item Centrality measures
    \item Local clustering coefficients
\end{itemize}

\subsection{Graphormer Integration}
We integrate TAPE embeddings with Graphormer through a multi-head attention mechanism incorporate spatial, centrality and edge encoding to capture the whole structural information of graph
\begin{equation}
    \text{Attention}(Q,K,V) = \text{softmax}(\frac{QK^T}{\sqrt{d_k}})V
\end{equation}
where $Q$, $K$, and $V$ are derived from node features and structural encodings. This mechanism:
\begin{itemize}
    \item In the attention mechanism, \textbf{node centrality} plays an important role as it indicates the importance of a node in a graph. To leverage this information, we introduce Centrality Encoding, which assigns each node two embedding vectors based on its indegree and outdegree. The centrality encodings are added to the node features as follows:
\[
h^{(0)}_i = x_i + z_{- \text{deg}(v_i)} + z_{+ \text{deg}(v_i)},
\]
where \( z_{-} \in \mathbb{R}^d \) and \( z_{+} \in \mathbb{R}^d \) are learnable embeddings for the indegree and outdegree of node \( v_i \), respectively. This encoding allows the model to capture both semantic and importance-related information during attention calculations.

\item The \textbf{Spatial Encoding} is designed to encode the structural relationships between nodes by considering their relative positions in the graph. For any pair of nodes \( v_i \) and \( v_j \), we define a function \( \varphi(v_i, v_j) \) to measure their spatial relationship, which in this case is the shortest path distance between them. The attention matrix is modified as follows:
\[
A_{ij} = \frac{(h_i W_Q)(h_j W_K)^T}{\sqrt{d}} + b_{\varphi(v_i, v_j)},
\]
where \( b_{\varphi(v_i, v_j)} \) is a learnable scalar depending on the shortest path distance between nodes \( v_i \) and \( v_j \). This allows the model to adaptively focus on nearby nodes while considering their structural relationship.

\item \textbf{Edge Encoding} is used to directly integrate edge features into the attention layers, ensuring that the relationships between nodes are influenced by the features of the edges connecting them. For each node pair \( (v_i, v_j) \), the average dot-product of the edge features along the shortest path between the nodes is computed and added to the attention matrix:
\[
A_{ij} = \frac{(h_i W_Q)(h_j W_K)^T}{\sqrt{d}} + b_{\varphi(v_i, v_j)} + c_{ij},
\]
where \( c_{ij} \) is computed as:
\[
c_{ij} = \frac{1}{N} \sum_{n=1}^N x_{e_n} (w^E_n)^T,
\]
where \( x_{e_n} \) is the feature of the \( n \)-th edge in the shortest path from \( v_i \) to \( v_j \), and \( w^E_n \) is the learnable weight embedding for the edge. This encoding ensures that edge features are utilized effectively in the attention mechanism.

\end{itemize}

\subsection{Node Classification}
For the final classification, we employ:

\paragraph{Loss Function} Cross-entropy loss with label smoothing:
\begin{equation}
    \mathcal{L} = -\sum_{i=1}^{N}\sum_{c=1}^{C}y_{ic}\log(\hat{y}_{ic})
\end{equation}
where $y_{ic}$ denotes the true class and $\hat{y}_{ic}$ is the predicted probability for class $c$.

\paragraph{Training Strategy} We implement:
\begin{itemize}
    \item Gradient accumulation for handling large graphs
    \item Learning rate warmup and decay
    \item Early stopping based on validation performance
\end{itemize}

This comprehensive methodology addresses our key challenges: semantic-structural integration through TAPE and Graphormer fusion, long-range dependencies via attention mechanisms, temporal evolution through LLM-based processing, and scalability through efficient architecture design.
\section{Dataset}
The \textit{ogbn-arxiv} dataset (a part of the Open Graph Benchmark (OGB)) is a large-scale directed citation network, where each node represents a research paper, and edges indicate citations. Each node is initially represented by a 128-dimensional feature vector derived from the \textbf{GloVe embeddings} of the paper's title and abstract. The chronological split of the dataset ensures that older papers are used for training and validation, while newer papers form the test set, mimicking real-world forecasting. We adopt the \textbf{TAPE (Text-Attributed Graph Representation Enhancement)} framework, which leverages \textbf{LLMs} such as GPT or LLaMA to extract semantic insights from textual metadata. To obtain TAPE-enhanced embeddings, the paper titles and abstracts are preprocessed and passed through an LLM to generate contextual explanations or embeddings. These outputs are then processed through a lightweight \textbf{LM interpreter} (e.g., a transformer) to produce compact, informative embeddings suitable for graph models. These enhanced embeddings either replace or augment the original 128-dimensional GloVe features, providing deeper semantic representation for each node. Integrating TAPE into the node classification pipeline allows for a more nuanced understanding of the dataset by combining both \textbf{textual insights} and \textbf{structural graph information}.

\section{Experimentation Setting}
To evaluate our model, we train and test on ogbn-arxiv dataset. Dataset contains 169,343 papers and 1,166,243 citattion. We consider a realistic data split based on the publication dates of the papers. The general setting is that the ML models are trained on existing papers and then used to predict the subject areas of newly-published papers. This supports the direct application of them into real-world scenarios, such as helping the arXiv moderators. Specifically, we train our model on papers published until 2017, validate on those published in 2018, and test on those published since 2019. We run our experimentation on NVIDIA A40 GPU. We use 100 epoch and 0.002 learning rate to train the model. It taks 40s to train the model. We evaluate our model using four well-known metrics: accuracy, precision, recall, and f1-score.

\section{Replication Package}
We share all the data and all the sources to replicate our study in \textcolor{blue}{\href{https://github.com/Nibir088/GML-Project-Node-Classification-}{GML-Project}}.

\section{Evaluation Metrics}

In evaluating the performance of the framework, the following metrics are commonly used for node classification:

\begin{enumerate}
    \item \textbf{Accuracy}: Accuracy measures the percentage of correctly classified nodes. If we use OGBarxiv, there'll be 40 classes.
\[
\text{Accuracy} = \frac{\text{Number of Correct Predictions}}{\text{Total Number of Predictions}}
\]

\item \textbf{Precision, Recall, and F1-score}:
Precision evaluates the proportion of correctly predicted positive instances out of all positive predictions
\[
\text{Precision} = \frac{\text{True Positives (TP)}}{\text{True Positives (TP)} + \text{False Positives (FP)}}
\]

\textbf{Recall} (also known as Sensitivity or True Positive Rate) measures the proportion of actual positives that were correctly identified:

\[
\text{Recall} = \frac{\text{True Positives (TP)}}{\text{True Positives (TP)} + \text{False Negatives (FN)}}
\]

\textbf{F1-score} is the harmonic mean of Precision and Recall, providing a balance between the two:

\[
\text{F1-score} = 2 \times \frac{\text{Precision} \times \text{Recall}}{\text{Precision} + \text{Recall}}
\]
Note that, we have 40 classes for this task. Therefore, we report the macro Precision, Recall, and F1 score as follows:
\begin{equation*}
    \text{macro Precision} = \frac{\sum_{i=1}^{i=N}\text{Precision}_i}{N}
\end{equation*}
\begin{equation*}
    \text{macro Recall} = \frac{\sum_{i=1}^{i=N}\text{Recall}_i}{N}
\end{equation*}
\begin{equation*}
    \text{macro F1} = \frac{\sum_{i=1}^{i=N}\text{F1}_i}{N}
\end{equation*}
where, $\text{Precision}_i$, $\text{Recall}_i$, and $\text{F1}_i$ denote Precision, Recall, and F1 score for $i^{th}$ class respectively. 




\end{enumerate}

These metrics provide a comprehensive view of the performance of the framework for node classification.

\section{Result Analysis}
\begin{table}[htbp]
    \centering
    \caption{Comparison of Model Performance Across Metrics}
    \label{tab:model_comparison}
    \begin{tabular}{lcccccc}
        \toprule
        Model & Accuracy & Precision & Recall & F1-Score\\
        \midrule
        GCN         & 0.713 & 0.558 & 0.492 & 0.508 \\
        MLP         & 0.539 & 0.356 & 0.318 & 0.322 \\
        SAGE        & 0.701 & 0.556 & 0.447 & 0.470 \\
        RevGAT      & 0.702 & 0.521 & 0.450 & 0.459 \\ \hline
        TAPE + GCN       & 0.744 & 0.623 & 0.542 & 0.561 \\
        TAPE + MLP       & 0.759 & 0.620 & 0.555 & 0.564 \\
        TAPE + SAGE       & 0.767 & 0.661 & 0.558 & 0.573 \\
        TAPE + RevGAT       & 0.761 & 0.603 & 0.570 & 0.590 \\
        \hline
        TAPE + Graphormer & 0.772 & 0.671 & 0.577 & 0.61\\
        (Our Approach) &&&&\\
        \bottomrule
    \end{tabular}
\end{table}

The results, as shown in Table \ref{tab:model_comparison}, indicate that the proposed TAPE + Graphormer framework outperforms baseline methods in all evaluated metrics. Specifically, the accuracy improved by nearly 10\% compared to the baseline (GCN), reflecting the effectiveness of integrating semantic and structural features. Precision and recall metrics also show marked improvement, suggesting a more balanced model that captures positive instances while minimizing false positives and negatives.

The combination of TAPE and Graphormer proves particularly beneficial in leveraging both textual and graph-based information. This integration enables the model to harness semantic richness and structural dependencies, resulting in a comprehensive understanding of node relationships. Additionally, the use of advanced positional encoding and attention mechanisms in Graphormer enhances the model's ability to capture long-range dependencies within the graph.

\section{Ablation Study}
An ablation study further validates the contributions of individual components, showing that both the TAPE and Graphormer modules independently contribute to improved performance. This indicates the robustness of the proposed framework and its potential applicability to other datasets with similar characteristics. Table \ref{ablation_table} presents the results of an ablation study, showcasing the impact of different components in the framework on validation and test accuracy.

\begin{table}[H]
\centering
\caption{Ablation Study: Impact of Framework Components}

\begin{tabular}{@{}lcc@{}}
\toprule
Configuration & Validation Accuracy & Test Accuracy \\
\midrule
Graphormer + TA & 0.7582 & 0.7530 \\
Graphormer + P & 0.7442 & 0.7382 \\
Graphormer + E & 0.0883 & 0.0881 \\
TA + P + E & 0.7307 & 0.7309 \\
\bottomrule
\end{tabular}
\label{ablation_table}
\end{table}



    
        
    

\section{Discussion}
The TAPE and Graphormer methodologies demonstrate advancements in enhancing node classification tasks within citation networks. By combining text analysis with graph structure, these approaches deliver a more nuanced understanding of scientific papers and their interrelationships. TAPE leverages large language models to generate semantically rich embeddings, effectively capturing the textual essence of research articles. Meanwhile, Graphormer enriches these embeddings by integrating positional and structural graph information, which aids in recognizing intricate citation patterns.

This framework is especially valuable for handling large-scale networks, offering potential benefits in streamlining research processes, improving recommendation systems, and enabling automated categorization of papers. For instance, moderators of platforms like arXiv could utilize these tools to classify submissions more efficiently.

However, despite its strengths, the framework's reliance on significant computational resources poses challenges for scalability and accessibility. Additionally, it currently lacks mechanisms to address the dynamic and evolving nature of citation networks, where new connections and nodes continuously emerge. These limitations highlight the need for ongoing optimization and adaptability in future iterations.

Future work will aim to overcome these challenges by optimizing the framework for efficiency, incorporating mechanisms to adapt to dynamic networks, and fostering transparency in its decision-making processes. Additionally, it will explore strategies to ensure fair and unbiased classification, emphasizing ethical responsibility in the application of such advanced technologies.

\section{Conclusion}
The integration of TAPE and Graphormer presents a robust and innovative framework for node classification in citation networks. By combining advanced text analysis with graph structural insights, the approach demonstrates superior performance in accurately categorizing nodes, as evidenced by significant improvements in accuracy and other evaluation metrics. While the framework addresses key challenges in leveraging textual and structural data, it also highlights limitations such as high computational demands and the need for adaptability to dynamic networks. Future efforts will focus on refining the framework to enhance efficiency, scalability, and ethical considerations, ensuring its broader applicability in research evaluation and related domains.

\bibliographystyle{ACM-Reference-Format}
\bibliography{main}


\begin{thebibliography}{18}


\ifx \showCODEN    \undefined \def \showCODEN     #1{\unskip}     \fi
\ifx \showDOI      \undefined \def \showDOI       #1{#1}\fi
\ifx \showISBNx    \undefined \def \showISBNx     #1{\unskip}     \fi
\ifx \showISBNxiii \undefined \def \showISBNxiii  #1{\unskip}     \fi
\ifx \showISSN     \undefined \def \showISSN      #1{\unskip}     \fi
\ifx \showLCCN     \undefined \def \showLCCN      #1{\unskip}     \fi
\ifx \shownote     \undefined \def \shownote      #1{#1}          \fi
\ifx \showarticletitle \undefined \def \showarticletitle #1{#1}   \fi
\ifx \showURL      \undefined \def \showURL       {\relax}        \fi
\providecommand\bibfield[2]{#2}
\providecommand\bibinfo[2]{#2}
\providecommand\natexlab[1]{#1}
\providecommand\showeprint[2][]{arXiv:#2}

\bibitem[Brown et~al\mbox{.}(2020)]%
        {brown2020language}
\bibfield{author}{\bibinfo{person}{Tom~B Brown}, \bibinfo{person}{Benjamin Mann}, \bibinfo{person}{Nick Ryder}, \bibinfo{person}{Melanie Subbiah}, \bibinfo{person}{Jared Kaplan}, \bibinfo{person}{Prafulla Dhariwal}, \bibinfo{person}{Arvind Neelakantan}, \bibinfo{person}{Pranav Shyam}, \bibinfo{person}{Girish Sastry}, \bibinfo{person}{Amanda Askell}, {et~al\mbox{.}}} \bibinfo{year}{2020}\natexlab{}.
\newblock \showarticletitle{Language Models are Few-Shot Learners}.
\newblock \bibinfo{journal}{\emph{Advances in Neural Information Processing Systems (NeurIPS)}} (\bibinfo{year}{2020}).
\newblock


\bibitem[Chiang et~al\mbox{.}(2019)]%
        {chiang2019cluster}
\bibfield{author}{\bibinfo{person}{Wei-Lin Chiang}, \bibinfo{person}{Xuanqing Liu}, \bibinfo{person}{Si Si}, \bibinfo{person}{Yang Li}, \bibinfo{person}{Samy Bengio}, {and} \bibinfo{person}{Cho-Jui Hsieh}.} \bibinfo{year}{2019}\natexlab{}.
\newblock \showarticletitle{Cluster-GCN: An Efficient Algorithm for Training Deep and Large Graph Convolutional Networks}. In \bibinfo{booktitle}{\emph{Proceedings of the 25th ACM SIGKDD International Conference on Knowledge Discovery \& Data Mining (KDD)}}.
\newblock


\bibitem[Chien et~al\mbox{.}(2021)]%
        {chien2021giant}
\bibfield{author}{\bibinfo{person}{Eli Chien}, \bibinfo{person}{Albert Gu}, \bibinfo{person}{Lihao Shen}, \bibinfo{person}{Aojun Zhang}, \bibinfo{person}{Jerry Li}, \bibinfo{person}{Christopher Zhang}, {and} \bibinfo{person}{Hung Nguyen}.} \bibinfo{year}{2021}\natexlab{}.
\newblock \showarticletitle{GIANT: Scalable Text-Attributed Graph Representation Learning}. In \bibinfo{booktitle}{\emph{Proceedings of the 37th International Conference on Machine Learning (ICML)}}.
\newblock


\bibitem[Chowdhery et~al\mbox{.}(2022)]%
        {chowdhery2022palm}
\bibfield{author}{\bibinfo{person}{Aakanksha Chowdhery} {et~al\mbox{.}}} \bibinfo{year}{2022}\natexlab{}.
\newblock \showarticletitle{PaLM: Scaling Language Modeling with Pathways}.
\newblock \bibinfo{journal}{\emph{arXiv preprint arXiv:2204.02311}} (\bibinfo{year}{2022}).
\newblock


\bibitem[Devlin et~al\mbox{.}(2019)]%
        {devlin2018bert}
\bibfield{author}{\bibinfo{person}{Jacob Devlin}, \bibinfo{person}{Ming-Wei Chang}, \bibinfo{person}{Kenton Lee}, {and} \bibinfo{person}{Kristina Toutanova}.} \bibinfo{year}{2019}\natexlab{}.
\newblock \showarticletitle{BERT: Pre-training of Deep Bidirectional Transformers for Language Understanding}. In \bibinfo{booktitle}{\emph{Proceedings of the 2019 Conference of the North American Chapter of the Association for Computational Linguistics (NAACL)}}.
\newblock


\bibitem[Harris(1954)]%
        {harris1954distributional}
\bibfield{author}{\bibinfo{person}{Zellig~S Harris}.} \bibinfo{year}{1954}\natexlab{}.
\newblock \showarticletitle{Distributional Structure}.
\newblock \bibinfo{journal}{\emph{Word}} \bibinfo{volume}{10}, \bibinfo{number}{2-3} (\bibinfo{year}{1954}), \bibinfo{pages}{146--162}.
\newblock


\bibitem[Hu et~al\mbox{.}(2020b)]%
        {hu2020open}
\bibfield{author}{\bibinfo{person}{Weihua Hu}, \bibinfo{person}{Matthias Fey}, \bibinfo{person}{Marinka Zitnik}, \bibinfo{person}{Yuxiao Dong}, \bibinfo{person}{John Willatt}, \bibinfo{person}{Bowen Liu}, \bibinfo{person}{Michele Catasta}, {and} \bibinfo{person}{Jure Leskovec}.} \bibinfo{year}{2020}\natexlab{b}.
\newblock \showarticletitle{Open Graph Benchmark: Datasets for Machine Learning on Graphs}. In \bibinfo{booktitle}{\emph{Proceedings of the 33rd Conference on Neural Information Processing Systems (NeurIPS)}}.
\newblock


\bibitem[Hu et~al\mbox{.}(2020a)]%
        {hu2020gpt}
\bibfield{author}{\bibinfo{person}{Ziniu Hu}, \bibinfo{person}{Yuxiao Dong}, \bibinfo{person}{Kuansan Wang}, \bibinfo{person}{Shiyu Chang}, {and} \bibinfo{person}{Yizhou Sun}.} \bibinfo{year}{2020}\natexlab{a}.
\newblock \showarticletitle{GPT-GNN: Generative Pre-Training of Graph Neural Networks}. In \bibinfo{booktitle}{\emph{Proceedings of the 26th ACM SIGKDD International Conference on Knowledge Discovery \& Data Mining (KDD)}}.
\newblock


\bibitem[Kipf and Welling(2016)]%
        {kipf2016semi}
\bibfield{author}{\bibinfo{person}{Thomas~N Kipf} {and} \bibinfo{person}{Max Welling}.} \bibinfo{year}{2016}\natexlab{}.
\newblock \showarticletitle{Semi-Supervised Classification with Graph Convolutional Networks}. In \bibinfo{booktitle}{\emph{Proceedings of the International Conference on Learning Representations (ICLR)}}.
\newblock


\bibitem[Mikolov et~al\mbox{.}(2013)]%
        {mikolov2013efficient}
\bibfield{author}{\bibinfo{person}{Tomas Mikolov}, \bibinfo{person}{Kai Chen}, \bibinfo{person}{Greg Corrado}, {and} \bibinfo{person}{Jeffrey Dean}.} \bibinfo{year}{2013}\natexlab{}.
\newblock \showarticletitle{Efficient Estimation of Word Representations in Vector Space}. In \bibinfo{booktitle}{\emph{Proceedings of the International Conference on Learning Representations (ICLR)}}.
\newblock


\bibitem[Touvron et~al\mbox{.}(2023)]%
        {touvron2023llama}
\bibfield{author}{\bibinfo{person}{Hugo Touvron} {et~al\mbox{.}}} \bibinfo{year}{2023}\natexlab{}.
\newblock \showarticletitle{LLaMA: Open and Efficient Foundation Language Models}.
\newblock \bibinfo{journal}{\emph{arXiv preprint arXiv:2302.13971}} (\bibinfo{year}{2023}).
\newblock


\bibitem[Velickovic et~al\mbox{.}(2018)]%
        {velickovic2017graph}
\bibfield{author}{\bibinfo{person}{Petar Velickovic}, \bibinfo{person}{Guillem Cucurull}, \bibinfo{person}{Arantxa Casanova}, \bibinfo{person}{Adriana Romero}, \bibinfo{person}{Pietro Li{\`o}}, {and} \bibinfo{person}{Yoshua Bengio}.} \bibinfo{year}{2018}\natexlab{}.
\newblock \showarticletitle{Graph Attention Networks}. In \bibinfo{booktitle}{\emph{Proceedings of the International Conference on Learning Representations (ICLR)}}.
\newblock


\bibitem[Velickovic et~al\mbox{.}(2019)]%
        {velickovic2019deep}
\bibfield{author}{\bibinfo{person}{Petar Velickovic}, \bibinfo{person}{William Fedus}, \bibinfo{person}{Will Hamilton}, \bibinfo{person}{Pietro Li{\`o}}, \bibinfo{person}{Yoshua Bengio}, {and} \bibinfo{person}{R~Devon Hjelm}.} \bibinfo{year}{2019}\natexlab{}.
\newblock \showarticletitle{Deep Graph Infomax}. In \bibinfo{booktitle}{\emph{Proceedings of the International Conference on Learning Representations (ICLR)}}.
\newblock


\bibitem[Yang et~al\mbox{.}(2016)]%
        {yang2016revisiting}
\bibfield{author}{\bibinfo{person}{Zhilin Yang}, \bibinfo{person}{William~W Cohen}, {and} \bibinfo{person}{Ruslan~R Salakhutdinov}.} \bibinfo{year}{2016}\natexlab{}.
\newblock \showarticletitle{Revisiting Semi-Supervised Learning with Graph Embeddings}. In \bibinfo{booktitle}{\emph{Proceedings of the 33rd International Conference on Machine Learning (ICML)}}.
\newblock


\bibitem[Yasunaga et~al\mbox{.}(2022)]%
        {yasunaga2022dragon}
\bibfield{author}{\bibinfo{person}{Michihiro Yasunaga}, \bibinfo{person}{Zhuoyuan Zhou}, \bibinfo{person}{Haoming Xu}, \bibinfo{person}{Mehran Zafarani}, \bibinfo{person}{Sebastian Riedel}, {and} \bibinfo{person}{Julian McAuley}.} \bibinfo{year}{2022}\natexlab{}.
\newblock \showarticletitle{DRAGON: Improving Graph Learning by Revisiting Feature Attribution}. In \bibinfo{booktitle}{\emph{Proceedings of the 36th Conference on Neural Information Processing Systems (NeurIPS)}}.
\newblock


\bibitem[Ying et~al\mbox{.}(2021)]%
        {ying2021transformers}
\bibfield{author}{\bibinfo{person}{Chengxuan Ying}, \bibinfo{person}{Tianle Cai}, \bibinfo{person}{Shengjie Luo}, \bibinfo{person}{Shuxin Zheng}, \bibinfo{person}{Guolin Ke}, \bibinfo{person}{Di He}, \bibinfo{person}{Yanming Shen}, {and} \bibinfo{person}{Tie-Yan Liu}.} \bibinfo{year}{2021}\natexlab{}.
\newblock \showarticletitle{Do Transformers Really Perform Bad for Graph Representation?}. In \bibinfo{booktitle}{\emph{Advances in Neural Information Processing Systems (NeurIPS)}}.
\newblock


\bibitem[Zhao et~al\mbox{.}(2022)]%
        {zhao2022gtn}
\bibfield{author}{\bibinfo{person}{Peihao Zhao}, \bibinfo{person}{Yizhen Liu}, \bibinfo{person}{Keyulu Xu}, \bibinfo{person}{Qifan Sun}, {and} \bibinfo{person}{Charles Zhang}.} \bibinfo{year}{2022}\natexlab{}.
\newblock \showarticletitle{GLEM: Global-to-Local Graph Neural Networks for Enhanced Graph Learning}. In \bibinfo{booktitle}{\emph{Proceedings of the 36th Conference on Neural Information Processing Systems (NeurIPS)}}.
\newblock


\bibitem[Zhu et~al\mbox{.}(2021)]%
        {zhu2021textgnn}
\bibfield{author}{\bibinfo{person}{Fengshun Zhu}, \bibinfo{person}{Fuli Feng}, \bibinfo{person}{Xiangnan Zhang}, \bibinfo{person}{Zhe Wan}, \bibinfo{person}{Xiang Wang}, \bibinfo{person}{Xiangnan He}, \bibinfo{person}{Yongdong Huang}, {and} \bibinfo{person}{Tat-Seng Chua}.} \bibinfo{year}{2021}\natexlab{}.
\newblock \showarticletitle{TextGNN: Improving Text Encoder via Graph Neural Network in Sponsored Search}. In \bibinfo{booktitle}{\emph{Proceedings of the 44th International ACM SIGIR Conference on Research and Development in Information Retrieval (SIGIR)}}.
\newblock


\end{thebibliography}
\end{document}